\documentclass[conference]{IEEEtran}
\IEEEoverridecommandlockouts

\usepackage[utf8]{inputenc} 
\usepackage[T1]{fontenc}    
\usepackage{hyperref}       
\usepackage{url}            
\usepackage{booktabs}       
\usepackage{amsfonts}       
\usepackage{nicefrac}       
\usepackage{microtype}      
\usepackage{algorithm,algpseudocode}
\usepackage{bbm}
\usepackage{hyperref}

\usepackage[table]{xcolor}

\usepackage{cite}
\usepackage{amsmath,amssymb}
\usepackage{graphicx}
\usepackage{textcomp}
\def\BibTeX{{\rm B\kern-.05em{\sc i\kern-.025em b}\kern-.08em
    T\kern-.1667em\lower.7ex\hbox{E}\kern-.125emX}}

\usepackage{amssymb}

\newcommand{\mbb}[1]{\mathbb{#1}}
\newcommand{\intCP}[2]{\mathcal{T}{(#1;#2)}}
\newcommand{\prob}[1]{\mathbb{P}{(#1)}}

\newcommand{\mcal}[1]{\mathcal{#1}}
\newcommand{\Z}{\mcal Z}

\newcommand{\I}{\mcal I}
\usepackage{amsmath}

\DeclareMathOperator*{\argmin}{arg\,min}

\usepackage{mathtools}

\newcommand{\smallQuad}{\hspace{0.1cm}}
\usepackage{lscape}
\usepackage{rotating}
\usepackage{amsmath}
\usepackage{amssymb}
\usepackage{mathtools}
\usepackage{amsthm}
\usepackage{multirow}
\usepackage{breqn}
\usepackage{caption}
\usepackage{subcaption}
\usepackage{tikz}
\usetikzlibrary{arrows, arrows.meta}
\usepackage{bbm}
\usepackage{adjustbox}
\usepackage[capitalize,noabbrev]{cleveref}
\theoremstyle{plain}
\newtheorem{theorem}{Theorem}[section]

\theoremstyle{definition}

\theoremstyle{remark}
\newtheorem{remark}[theorem]{Remark}

\newcommand{\mb}[1]{\mathbf{#1}}
\usepackage[textsize=tiny]{todonotes}

\begin{document}

\title{An In-Depth Examination of Risk Assessment in Multi-Class Classification Algorithms
}

\author{
\IEEEauthorblockN{Disha Ghandwani}
\IEEEauthorblockA{\textit{Stanford University} \\
California, US \\
disha123@stanford.edu}
\and
\IEEEauthorblockN{Neeraj Sarna}
\IEEEauthorblockA{\textit{Munich RE} \\
Munich, Germany \\
nsarna@munichre.com}
\and
\IEEEauthorblockN{Yuanyuan Li}
\IEEEauthorblockA{\textit{Munich RE} \\
California, US  \\
yli@munichre.com}
\and
\IEEEauthorblockN{Yang Lin}
\IEEEauthorblockA{\textit{Hartford Steam Boiler} \\
Connecticut, US \\
nsarna@munichre.com}
}

\maketitle

\begin{abstract}
Advanced classification algorithms are being increasingly used in safety-critical applications like health-care, engineering, etc. In such applications, miss-classifications made by ML algorithms can result in substantial financial or health-related losses. To better anticipate and prepare for such losses, the algorithm user seeks an estimate for the probability that the algorithm miss-classifies a sample. We refer to this task as the risk-assessment.
For a variety of models and datasets, we numerically analyze the performance of different methods in solving the risk-assessment problem. We consider two solution strategies: a) calibration techniques that calibrate the output probabilities of classification models to provide accurate probability outputs; and b) a novel approach based upon the prediction interval generation technique of conformal prediction. Our conformal prediction based approach is model and data-distribution agnostic, simple to implement, and provides reasonable results for a variety of use-cases. We compare the different methods on a broad variety of models and datasets. 
\end{abstract}


\section{Introduction}
\label{intro}

Advanced machine learning classification algorithms are being used in several safety-critical applications. Consider, for instance, melanoma detection using computer vision \cite{melanoma_cv,melanoma_cv2,melanoma_cv3}. In this case, a miss-classification from the model has dreadful consequences. A miss-classified healthy patient might be given a treatment it didn't need or worse, a miss-classified sick patient might not receive any immediate treatment at all. Another example is from fault detection of off-shore wind farms where a missed fault or a false alarm result in grave financial consequences \cite{PM_wind,PM_wind2}. In such cases, the model user would like to be well-prepared for model errors/miss-classifications. Thus, the question of interest is: what is the probability that the model miss-classifies a sample? We refer to this problem as risk-assessment.

We state the risk-assessment problem concretely. With $\mb{X} \in \mbb R^d$, $Y\in \{0,1,\dots,K\}$ and $\hat Y(X)\in \{0,1,\dots,K\}$ being the input, the true label and the model's output, respectively, we seek an estimate for the miss-classification probability $\prob{Y \not = \hat Y(X)}$. Note that $K$ represents the number of classes and for simplicity, we considered a point prediction $\hat Y(X)$. A model could also output a prediction interval (PI) containing the top-k classes. The expression for the miss-classification problem would then contain the PI instead of $\hat Y(X)$.

\textbf{Existing Methodologies:} Classification models provide a potential solution for risk-assessment. For a given input $X = x$, these models output the probability of the true label being in the different classes i.e., they estimate $\prob{Y = k | X = x}$. This estimation is often over-confident for the top classes i.e., it could be much higher than its true value \cite{tempscale,zadrozny2001obtaining,zadrozny2002transforming}. This could gravely underestimate the risk of model failure and is detrimental for safety-critical applications where---given the severity of consequences resulting from a miss-classification---overestimating the risk and thus being a little over-prepared for adversities is preferable. 

The probabilities from a classification model could nonetheless be calibrated to better reflect the true probabilities \cite{tempscale,zadrozny2001obtaining,zadrozny2002transforming,BBQ_binning}. The calibrated probabilities then provide a solution to the risk-assessment problem, which one would expect to be more accurate than using the raw model's probabilities. Using numerical experiments, we later compare the performance of various calibration techniques in solving the risk-assessment problem.

 \textbf{Conformal prediction approach:} In addition to comparing various calibration techniques, we propose a novel risk-assessment technique based upon conformal prediction (CP). CP generates PIs that contain the true label with a given coverage level, and is applicable to both regression (see \cite{conformal_pred, conformal_pred_intro, cqr,image_to_image_regression,bio_cp_regression,disentagled_latent_space}) and classification \cite{conformal_pred, conformal_pred_intro, cp_classficiation_image, cp_classficiation_image_2,cp_ordinal_classification,improve_expert_prediction}. The methodology relies on quantifying the model's uncertainty on a hold-out set via a score-function. This uncertainty is then used to construct PIs for new datapoints \cite{conformal_pred}. The choice of the score-function is paramount to CPs performance \cite{conformal_pred}. For classification, authors have developed several different score functions, each of which provide varying degree of accuracy \cite{classification_avg_size, cp_classficiation_image, cp_classficiation_image_2, conformal_pred_intro}.
 
In CP, the coverage level is given and the resulting PI contains the true label with at least this coverage level. To utilize it for risk-assessment, we turn this idea around---see \cite{JAWS,singh2023} for a similar approach for regression. Interpreting the model's output as a PI, we seek an answer to: what is the probability that the true label lies inside this PI? Notice that this reformulation is the same as the risk-assessment problem we defined earlier. This inspires the following approach to risk-assessment: a) approximate the model's output via a CP interval; and b) use the confidence level of the CP interval as a solution to the risk--assessment problem. To the best of our knowledge, ours is the first work that uses CP for quantifying the confidence of a classification model. Our work thereby extends the framework proposed by the authors in \cite{JAWS} to classification.

The conformal prediction approach falls into the category of non-parametric methods and offers the following benefits. Firstly, it is model agnostic and therefore, could be applied to any model that outputs probabilities for different classes. Secondly, being a non-parametric method, it doesn't make any assumption on the functional form of the true class probabilities. Lastly, it does not require any loss-function minimization and is therefore simpler to numerically implemented compared to the other calibration techniques.

\section{Calibration of classification models}
We present the idea behind calibration and how it solves the risk-assessment problem. Recall that $\mb{X}$, $\hat{Y}(\mb X)$ and $Y$ represent
the model input, model output and the true label, respectively. Additionally, let $\hat p$ represent the confidence of the model in its output i.e., the probability of correctness, which the model believes it has. For a perfectly calibrated model, the model's confidence $\hat p$ equals the true probability of correctness $p$. This means that for all $p\in [0,1]$, we would have 
\begin{gather}
\prob{Y = \hat Y(\mb X)| \hat p = p} = p.
\end{gather}
In other words, if we claim that the model is correct with a probability $p$ then, it is indeed correct with a probability $p$. We refer the reader to \cite{review_calibration} for additional details.

We recover a solution to risk-assessment as follows.
For a given input $\mb X$, calibration methods calibrate the probability $\hat p(\mb X)$ to $r(\mb X)$. To estimate $\prob{Y\not = \hat Y(\mb X)}$, we average $1-r(\mb X)$ over a representative input set---which could come from a test set, or another hold-out set set---$\Z^* := \{(\mb X_{i_*})\}_{i_*=1,\dots,m}$ to find
\begin{gather}
    \prob{Y \not = \hat Y(\mb X)} \approx \hat \alpha\quad \text{where}\quad \hat\alpha:= \frac{1}{|\Z^*|}\sum_{\mb X\in\Z^*}\left(1-r(\mb X)\right). \label{sum test set}
\end{gather}

A value of the calibrated probability $r(\mb X)$ could result from any of the different calibration technique. The three most common techniques that we use are discussed below.

\begin{remark}[ECE metric]
The quality of calibration is judged by the Expected Calibration Error (ECE) given by $ECE= E_{\hat p}|\prob{Y = \hat Y(\mb X)| \hat p = p} - p|$. The metric we would later (see \ref{sec:experiments}) use to access the quality of risk-assessment is the difference between the true miss-classification probability $ \prob{Y \not = \hat Y(\mb X)}$ and its approximation $\hat p$ (or equivalently $\hat \alpha$). Our metric targets the marginal failure probability $ \prob{Y \not = \hat Y(\mb X)}$ rather than bin based approach in ECE. Using Jensen's inequality, one could show that ECE is bounded from below by our metric. Therefore, the various studies related to calibration performed in the past (see \cite{tempscale,zadrozny2001obtaining,zadrozny2002transforming,review_calibration}) do not directly translate to our use-case, prompting the current investigation. 
\end{remark}

\subsection{Calibration Techniques}\label{sec:calibration}
This section discusses the application of existing calibration techniques to determine the calibration probability $r(\mb X)$ , which facilitates the estimation of $\hat{\alpha}$ given in \ref{sum test set}. For simplicity, only for this sub-section, we restrict to binary classification where the true label $Y\in \{0,1\}$--extension to multi-class classification follows from the one-versus-all approach \cite{zadrozny2001obtaining}. For a given input $\mb X$, we assume we have the model's output probabilities $\hat p(\mb X)$ for class $1$. Furthermore, for calibration, we consider a hold-out denoted by 
\begin{gather}
\mcal Z = \{(\mb X_{i},Y_i)\}_{i = 1,\dots, n}. \label{calibration set}
\end{gather}
This set is not used for training the model.

\textbf{Histogram binning} The probability domain $[0,1]$ is divided into non-overlapping bins $B_1,\dots,B_M$. The value of $M$ is user-defined and the bins could either be equally spaced or defined such that they contain the same number of points. Each bin gets a calibrated probability $\theta_m$ that is computed using the calibration set $\mcal Z$ via
\begin{gather}
    \min_{\theta_1,\dots,
    \theta_M} \sum_{j=1}^M\sum_{i = 1}^n \mathbbm{1}(a_j\leq \hat p(\mb X_i)\leq a_{j+1})(\theta_j - Y_i)^2 \label{hist bin}
\end{gather}
where $a_i$'s are the boundaries of the bins---see \cite{zadrozny2001obtaining, tempscale} for further details. For any input $\mb X=x$, the following steps result in the calibrated probability: a) obtain the output probability $\hat p(\mb X)$; b) find the bin $B_j$ in which $\hat p(\mb X)$ belongs; and c) the calibrated probability is the calibrated probability corresponding to the bin i.e., $r(\mb X) = \theta_j$. 

\textbf{Isotonic regression} generalizes histogram binning \cite{zadrozny2002transforming}. The calibrated probabilities result from applying a piecewise constant function $f$ to $\hat p(\mb X)$ i.e., $r(\mb X) = f(\hat p(\mb X))$. The function $f$ results from solving an optimization problem, which is the same as \eqref{hist bin} but with the bin boundaries being additional parameters. The optimization problem is solved over the calibration set $\Z$ and reads
\begin{gather}
    \min_{M,\theta_i,a_i}\sum_{j=1}^M\sum_{i = 1}^n \mathbbm{1}(a_j\leq \hat p(\mb X_i)\leq a_{j+1})(\theta_j - Y_i)^2,
\end{gather}
with the constraints that $0 = a_1\leq a_2 \leq \dots \leq a_{M+1} = 1$ and $\theta_1\leq \theta_2\dots\leq \theta_M$.
Note that such a formulation is possible because $f$ is piecewise constant. 

\textbf{Temperature scaling }is a simplication of Platt scaling \cite{tempscale}. Although there are various similar scaling techniques available \cite{nixon2019measuring}, we concentrate on temperature scaling.  We assume that the logits $z(\mb X)\in\mbb{R}$ are such that the (un-calibrated) output probability $\hat p(\mb X)$ could be recovered from $z(\mb X)$ by applying sigmoid activation function. The calibrated probability results from scaling the logits with a temperature parameter $T$ i.e., $r(\mb X) = \mathrm{sigmoid}(z(\mb X)/T)$, where $r(\mb X)$ and $z(\mb X)$ are the calibrated probability and logits, respectively. The temperature $T$ results from minimising the Negative Log-Likelihood (NLL) over the calibration set $\Z$. The class that has the maximum calibrated probability is then the model's output. Since temperature scaling does not change the maximum of the sigmoid activation, the model's output remains the same as pre-calibration. Usually, $T \geq 1$ resulting in a smearing out of the probability densities. As $T\to\infty$, the probability densities approach a uniform distribution over the different classes. Furthermore, for $T\to 0$, the density approaches to a point mass---see \cite{nixon2019measuring,tempscale} for further details. 

\section{Conformal Prediction (CP) Approach}
We start with an overview of the CP technique and then leverage it for risk-assessment. We built upon the previous works that develop a similar approach for regression---see \cite{JAWS} and its extension \cite{singh2023}. 
\subsection{Overview of CP}
Let $\alpha \in [0,1]$ be a given miss-coverage level. CP generates a prediction interval $\intCP{\mb X}{\alpha}$ such that the true label lies inside with a probability greater than $1-\alpha$ i.e.,
\begin{gather}
    \prob{Y\in \intCP{\mb X}{\alpha}} \geq 1-\alpha. \label{coverage CP}
\end{gather}
Under the assumption that the data is \textit{iid} (could be relaxed to exchangeability \cite{conformal_pred_intro}), the above coverage can be achieved by the following three steps \cite{conformal_pred}: (i) define a score function $s(x,y)\in \mbb R$ that quantifies the uncertainty of the model---exact form given below; (ii) compute the $\frac{\lceil (1 + n)  *(1-\alpha)\rceil}{n}$--th quantile of the scores $\{s_i = s(\mb{X}_i,Y_i)\}_{i=1,\dots,n}$, where $(\mb X_i,Y_i)$ are the calibration points defined in \eqref{calibration set}. Denote this quantile by $\hat q$; and (iii) for a new test point $\mb X = x_{n+1}$, compute the PI by collecting ouputs that have a score less than $\hat q$ i.e.,
\begin{gather}
\intCP{\mb X}{\alpha}= \{y\smallQuad : \smallQuad s(\mb X,y) \leq \hat q\}. \label{cp int}
\end{gather}

We refer to \cite{cp_classficiation_image} for proofs on why the above construction results in the coverage property given in \eqref{coverage CP}. Note that this is the so-called split version of CP \cite{split_cp_1}. Additional statistical efficiency can be obtained via a cross-validation type approach \cite{covariate_shift_1}. For its computational efficiency and simplicity, we stick to the split-CP approach. 

The score function $s(x,y)$ is defined as follows. We consider $K$-classes. The output probability of the model $\mb{\hat p}(\mb{X})$ is a vector of size $K$, and the model's output $\hat Y(\mb X)$ is the class that has the maximum probability. We consider a score function from the Adaptive Prediction Sets (APS) technique given as \cite{adapt_1, valid_adapt}
\begin{gather}
s(x,y) = \sum_{i = 1 }^k \mb{\hat p}(x)_{\pi_i(x)}, \smallQuad \text{where} \smallQuad y = \pi_k(x). \label{def:score}
\end{gather}
Here, $\{\pi_1(x),\dots,\pi_K(x)\}$ permutes the set $\{1,\dots,K\}$ such that the probabilities $\{\mb{\hat p}_{\pi_1(x)}(x),\dots,\mb{\hat p}_{\pi_K(x)}(x)\}$ have a descending order. This also implies that $\hat Y(\mb X) = \pi_1(\mb X)$. Empirically, we found that the Least Ambiguous Set-valued Classifier (LAC) score function---proposed in \cite{classification_avg_size}---provided inferior results and hence, we didn't explore it further. Recall that the LAC score function considers 1 - $\mb{\hat p}_{Y}(x)$ as the score, where $Y$ is the true label. 

\subsection{Inverse CP (InvCP) based solution to risk-assemsment} \label{sec: invcp}
Building upon \cite{JAWS}, the following discussion clarifies that the risk-assessment framework is the inverse of that of CP. We therefore refer to our approach as the InvCP approach. We first note that $\prob{Y\not = {\hat Y(\mb X)}} = \prob{Y\not \in \I(\mb X)}$, where the prediction interval $\I(\mb X) := \{\hat Y(\mb X)\}.$\footnote{Note that a model's output could also be the top-k classes with $k>1$. In that case, the interval $\I(\mb X)$ could be extended to include these classes. The below framework already accommodates for this extension.} The quantity $\prob{Y\not \in \I(\mb X)}$ is the miss-coverage level of the prediction interval. Owing to the above equality, an estimate for the miss-coverage level is also a valid estimate for the miss-classification probability. The current situation is thus contrary to that of CP, we know the PI (which $\I(\mb X)$) but not its miss-coverage level. 

Our problem then reduces to: How to estimate the miss-coverage level for the interval $\I(\mb X)$? We approximate this miss-coverage by the miss-coverage level of a CP interval---i.e., approximating $\I(\mb X)$ by $T(\mb X; \alpha(\mb X))$---that has the following two properties:
\begin{align}
    \text{(P1):}\smallQuad \text{contains $\I(X)$},\quad\text{(P2):}\smallQuad  \text{has the smallest size}.
\end{align}

\ref{remark:conservativeness} elaborates more on the above two properties. Precisely, for an input $X=x$, we find a miss-coverage level $\alpha(\mb X)$ such that 
\begin{gather}
\alpha(\mb X) := \argmin_{\alpha' \in [0,1]}\{|\intCP{\mb X}{\alpha'}|\smallQuad :\smallQuad \I(\mb X)\subseteq \intCP{\mb X}{\alpha'}\}, \label{def: alphX}
\end{gather}
    where $|\cdot|$ represents the size of a set. Averaging $\alpha(\mb X)$ over a test set $\Z^*$ (given in \eqref{sum test set}), we find
    \begin{gather}
    \prob{Y \not \in \I(\mb X)} \approx\hat \alpha \quad \text{where}\quad \hat\alpha = \frac{1}{|\Z^*|}\sum_{\mb X\in\Z^*}\alpha(\mb X).
    \end{gather}
Below, we present an algorithm to compute $\alpha(\mb X)$.

\begin{remark}[Relation to regression]\label{remark:conservativeness}
For regression, authors in \cite{JAWS,singh2023} compute a CP-interval that is smaller than $\I(\mb X)$. Since an interval that is smaller than $\I(\mb X)$ has a higher miss-coverage than $\I(\mb X)$, the goal in these prior works was provably conservative risk-assessment i.e., the computed failure probability should not be smaller than the true failure probability. Applying the same methology to classification results in an over-conservative risk-assessment, which is usually inaccurate. Therefore, in (P1), being slightly non-conservative, we consider an interval that is larger than $\I(X)$ but still of the smallest possible size. As presented later empirically, this methodology still provides conservative results while being accurate. Contrary to regression, since we are considering an interval larger than $\I(X)$, it is unclear as of yet if the conservativess can be guaranteed theoretically.
\end{remark}

\subsubsection{Computation of $\alpha(\mb X)$}
\ref{alg: invcp} summarizes the algorithm for computing $\alpha(\mb X)$. The details are as follows. We assume that the scores $\{s_i\}_{i=1,\dots,n}$ are sorted $s_1\leq s_2\leq \dots \leq s_n$.
Furthermore, for a given input $\mb X=x$, we compute the score $s(\mb X,\I(\mb X))$ for the interval $\I(\mb X)$ defined as $s(\mb X,\I(\mb X)) := \max \{s(\mb X,y)\}_{y\in \I(\mb X)}.$
In other words, after arranging the output probabilities of different classes in a decreasing order, we add all the probabilities till the last label in $\I(X)$.

Recall that we seek an $\alpha(\mb X)$ such that the interval $\intCP{\mb X}{\alpha(\mb X)}$ satisfies the properties (P1) and (P2) given above. Recalling the definition of a CP-interval \eqref{cp int}, we find that $
\intCP{\mb X}{\alpha(\mb X)}= \{y\smallQuad : \smallQuad s(\mb X,y) \leq \hat q(\mb X)\},$
where $\hat q(\mb X)$ is the $\frac{\lceil (1 + n)  *(1-\alpha(\mb X))\rceil}{n}$--th quantile of the calibration scores $\{s_i\}_{i=1,\dots,n}$ at point $\mb X=x$. Thus, finding an $\alpha(\mb X)$ is equivalent to finding a quantile level $\hat q(\mb X)$.

To satisfy the inclusion property in (P1), $\hat q(\mb X)$ must not be smaller than $s(\mb X,\I(\mb X))$ and to satisfy the size constraint in (P2), it should be the smallest possible. This implies that out of all the calibration scores $\{s_i\}_i$, we set
\begin{gather}
\hat q(\mb X) = s_{\gamma(\mb X)},\smallQuad \text{where}\smallQuad \gamma(\mb X) := \min\{i\smallQuad :\smallQuad s_i\geq s(\mb X,\I(\mb X))\}.\label{sol:alpha_x}
\end{gather}
The definition of a quantile then provides $\alpha(\mb X) = 1 - \frac{\gamma(\mb X)}{n+1}$.

\begin{remark}[Complexity Analysis]
Computing the calibration scores on the calibration set and sorting them requires $O(nK + n log(n))$ operations. During inference, for each input in a test set $\Z^*$ of size $m$, computing $s(\mb X,\I(\mb X))$ requires $O(1)$ operations and then computing $\gamma(\mb X)$ requires $O(n)$ operations. Consequently, the complexity of the entire InvCP algorithm is $O(nK+ n\log(n)+nm)$. Note that unlike other calibration techniques given in \ref{sec:calibration}, InvCP does not need convex optimizations or hyper-parameter. The implementation is thus easier and the algorithm is generally more efficient.

\end{remark}
\begin{algorithm}
\caption{InvCP for classification}\label{alg: invcp}
 \hspace*{\algorithmicindent} \textbf{Input:} calibration set $\Z$, test set $\Z^*$, set of model outputs  $\{\I(\mb X)\}_{\mb X\in\Z^*}$ \\
 \hspace*{\algorithmicindent} \textbf{Output:} approximation to miss-classification probability $\hat \alpha$
\begin{algorithmic}
\For{$(\mb X_i,Y_i)$ in $\Z$}
\State Compute $s_i = s(\mb X_i,Y_i)$
\EndFor
\State $\{s_i\}_i\leftarrow \text{sorted}\left(\{s_i\}_i\right)$
\For{$\mb X^*_i$ in $\Z^*$}
\State $s^*_i\leftarrow s(\mb X^*_i,\I(\mb X_i^*))$
\State $\gamma(\mb X^*_i)\leftarrow \min\{k\smallQuad : \smallQuad s_i^*\leq s_k\}$
\State $\alpha(\mb X^*_i)\leftarrow 1 - \gamma(\mb X^*_i)/(|\Z|+1)$
\EndFor
\State $\hat \alpha \leftarrow \frac{1}{|\Z^*|}\sum_{\mb X^*_i\in\Z^*} \alpha(\mb X^*_i) $
\end{algorithmic}
\end{algorithm}

\section{Experiments}\label{sec:experiments}
This section compares the performance of the various risk-assessment techniques for different models and datasets.

\textbf{Risk-assessment techniques: }We consider the following 6 techniques for risk-assessment. The SMX technique that directly uses a model's softmax outputs to solve risk-assessment. The PLATT, HIST-BIN and ISO-REG techniques are based upon Platt scaling, histogram binning and isotonic regression, respectively. The InvCP technique considers the CP-based approach outlined in \ref{sec: invcp}.

\textbf{Datasets: }We consider five datasets: (i) CIFAR-100 \cite{cifar100}; (ii) CIFAR-10; (iii) Flowers102 \cite{Flowers102}; (iv) ImageNet V1 \cite{deng2009imagenet}; and (v) Places 365 \cite{zhou2017places}. The validation set of each dataset is randomly split into a calibration set $\Z$ and a test set $\Z^*$. The size of the calibration set is $20\%$ of the validation set size. \ref{table: sizes datasets} summarizes the sizes of the various datasets and the corresponding splits. 

\begin{table}[H]
\centering
\begin{tabular}{ |c|c|c| } 
 \hline
 Dataset & $|\Z|$ & $|\Z^*|$ \\ 
 \hline
 CIFAR100 & 2000 & 8000 \\ 
 CIFAR10 & 2000 & 8000 \\ 
 Flowers102 & 1225 & 4905 \\ 
 ImageNet--V1 & 10000 & 40000 \\
 Places365 & 7300 & 29200\\
 \hline
\end{tabular}
\caption{Splits for different datasets}
\label{table: sizes datasets}
\end{table}

\textbf{Model architecture:} We consider both tree-based and convolutional neural network architecture. For the former, we consider LightGBM, AdaBoost, XGBoost and RandomForest. For the latter we consider resNets \cite{he2016deep}, denseNets \cite{huang2017densely}, AlexNet \cite{krizhevsky2017imagenet}, and VGGs \cite{simonyan2014very}. For CIFAR-100, CIFAR-10 and Flowers102 we pre-trained all the models. For ImageNet and Places365, we used pre-trained models available from \cite{paszke2019pytorch} and \cite{zhou2017places}, respectively. Due to computational complexity, we do not train tree-based models from scratch for these datasets. In \ref{training for image} we summarize the accuracy of various models considered. 

\textbf{Performance Metrics: }The true miss-classification rate is unavailable in practise. Our reference is therefore a counting-based empirical estimate given as $\alpha_{\text{Emp}} := \sum_{X\in\Z^*}\mathbbm{1} (Y\not \in \I(X))/|\Z^*|.$ We ensure that $\Z^*$ is large enough so that $\alpha_{\text{Emp}}$ provides a good approximation to the true miss-classification rate. We are interested in the deviation of $\hat \alpha$ from this empirical true value defined as 
\begin{gather}
\delta := \hat \alpha - \alpha_{\text{Emp}}. \label{def: delta}
\end{gather}
It is also desirable that in safety critical applications, we do not under-estimate the risk \cite{JAWS,singh2023}. For such use cases, it is better to be over than under-prepared. Equivalently, it is desirable to have 
\begin{gather}
    \text{Conservativeness: }\delta \geq 0.
\end{gather}
If the above is satisfied, we say that a risk-assessment method is \textit{conservative}. All of our results present the average of $\delta$ over 100 independent splits of calibariation and test set.
\subsection{Results}

\begin{figure}[htbp]
\centering
\begin{subfigure}[b]{0.45\textwidth}
\centering
         \includegraphics[width=6cm]{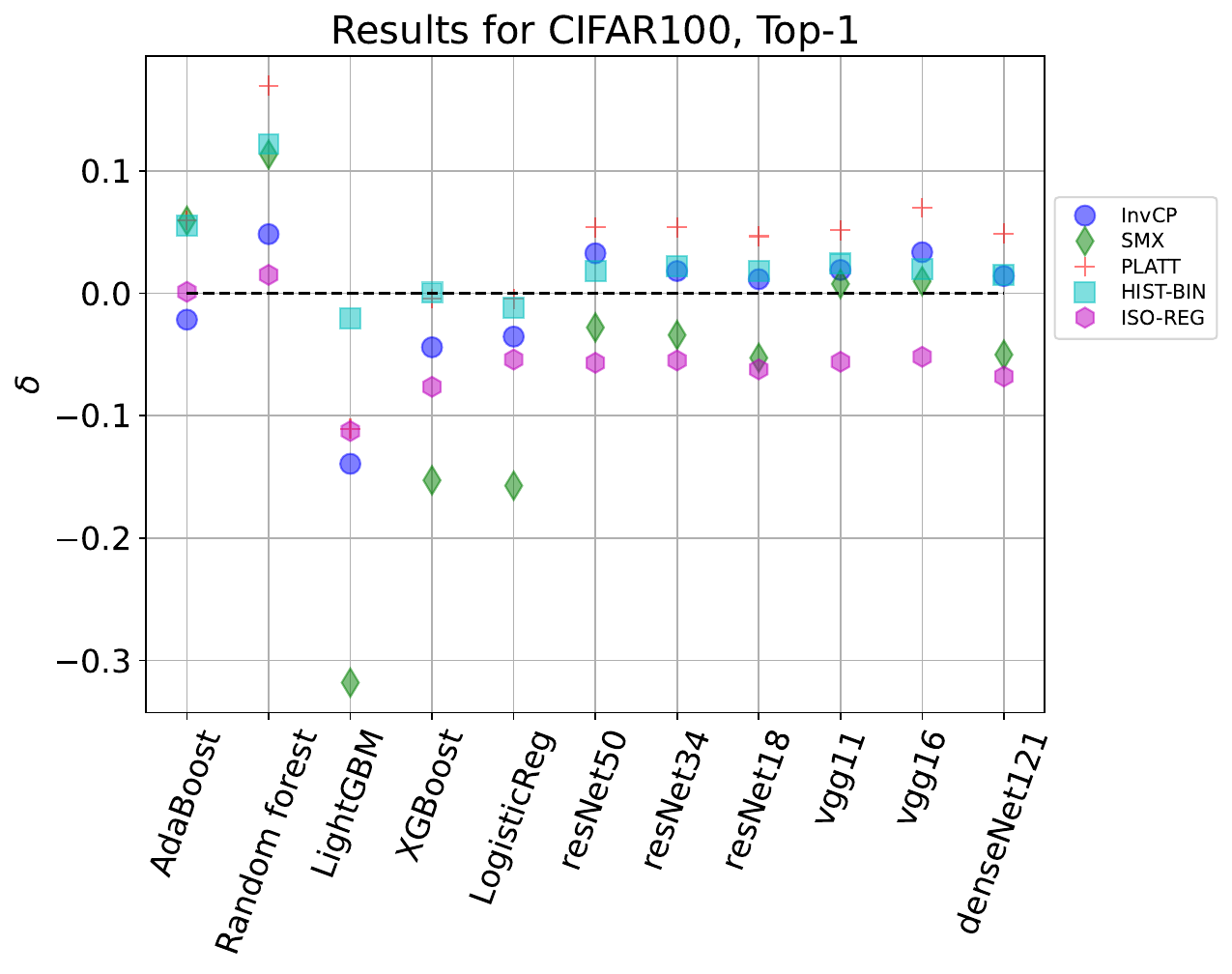}
         \caption{Top-1 results}
         \label{CIFAR100_1}
\end{subfigure}
\begin{subfigure}[b]{0.45\textwidth}
\centering
         \includegraphics[width=6cm]{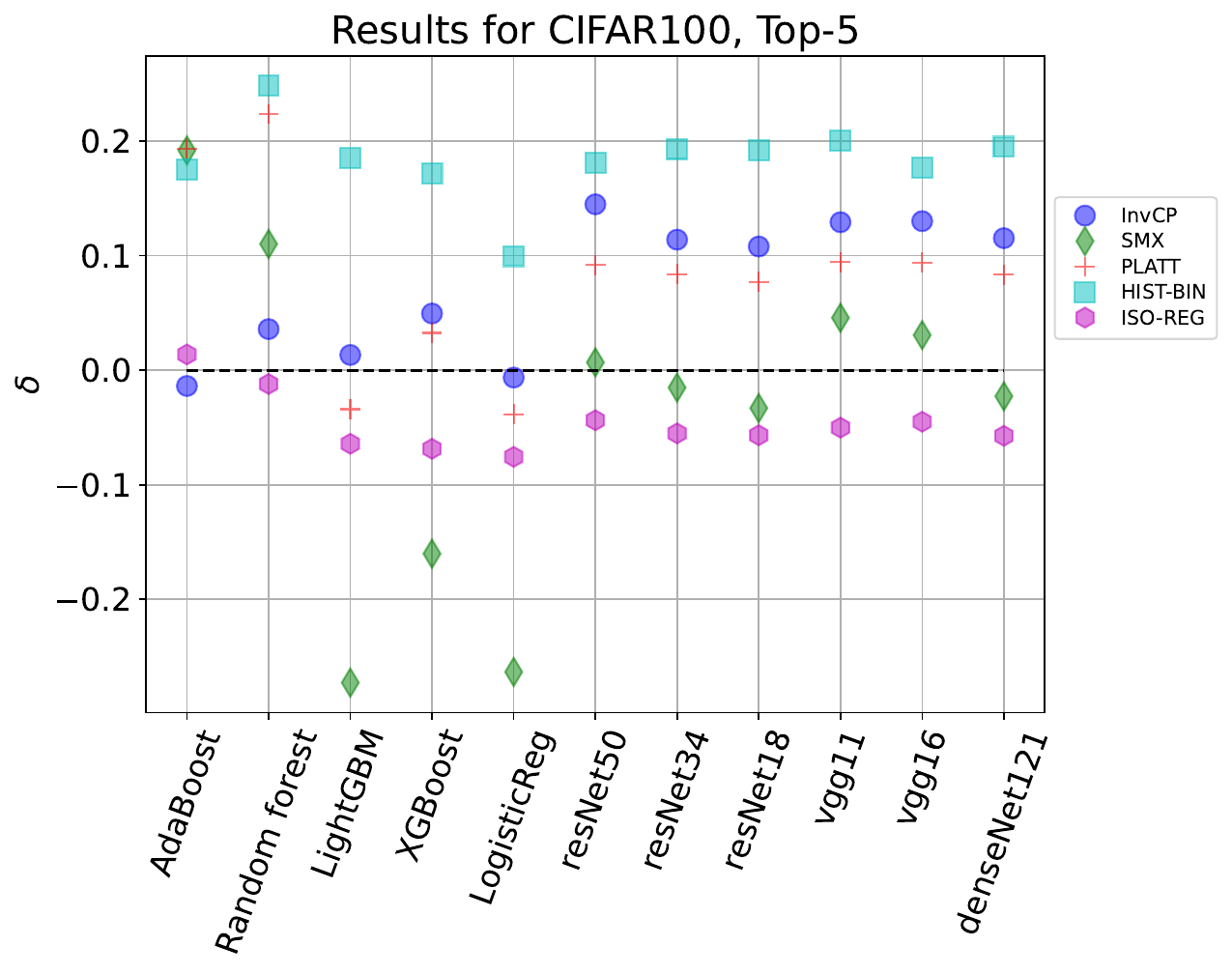}
         \caption{Top-5 results}
         \label{CIFAR100_5}
\end{subfigure}
  { \caption{Results for CIFAR-100. Smaller the $\delta$ (see \ref{def: delta}), more accurate the method. $\delta \geq 0$ implies conservativeness of the method. }}
\end{figure}

\textbf{CIFAR-100:} Results for top-1 and top-5 model outputs are shown in \ref{CIFAR100_1} and \ref{CIFAR100_5}, respectively. We consider two different model types, tree-based models and CNNs. For the top-1 prediction, InvCP provide results that are comparable to the standard calibration techniques. For all the CNNs, InvCP performs slightly better than PLATT, which, for CNNs, is usually the best performing calibration technique \cite{tempscale}. For both VGG11 and VGG16 models, SMX already provides the best results---none of the other techniques could improve upon this result. The results for HIST-BIN and InvCP are nonetheless very close to those of SMX. For the LightGBM model, compared to other tree models, InvCP provides worse results. Both the HIST-BIN and the InvCP method maintain conservativeness for most of the models. Despite the accuracy, ISO-REG is mostly not conservative. 

For the top-5 model output, the CNNs already provide good results with the plain SMX method. This is as expected. The top-5 accuracy of these models for CIFAR100 is high i.e., $1-\alpha_{\text{Emp}}$ is high. As a result, the SMX method that usually places high probability mass on the top classes provides results that are close to $1-\alpha_{\text{Emp}}$, leading to a small $\delta$. Observe that none of the other techniques could improve upon the SMX results. This is contrary to the tree-based models, where the SMX results are unreliable. For these models, both InvCP and ISO-REG provide accurate results. For these models, HIST-BIN, which performs well for the top-1 case, fails to provide accurate results here.

\textbf{CIFAR-10 and Flowers-102:} Results for both these datasets are very similar to those for CIFAR-100. One single risk-assessment method doesn't outperform all the other methods for all the models. Nonetheless, InvCP provides the best results in general while being free of optimization and hyper-parameter tuning---see \eqref{app: further results} for further details. 

\begin{figure}[htbp]
\centering
\begin{subfigure}[b]{0.45\textwidth}
\centering
         \includegraphics[width=6cm]{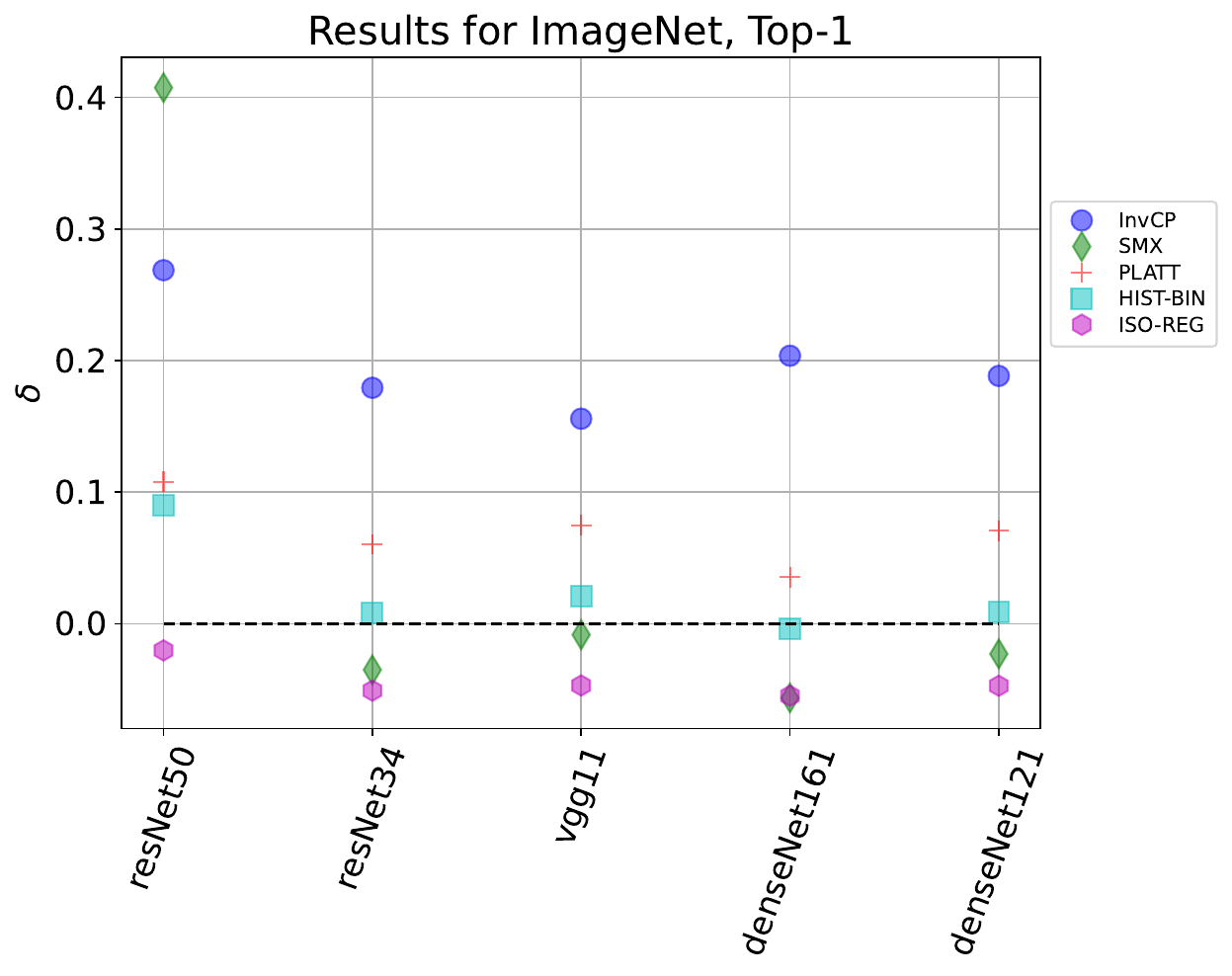}
         \caption{Top-1 results}
         \label{ImageNet_1}
\end{subfigure}
\begin{subfigure}[b]{0.45\textwidth}
\centering
         \includegraphics[width=6cm]{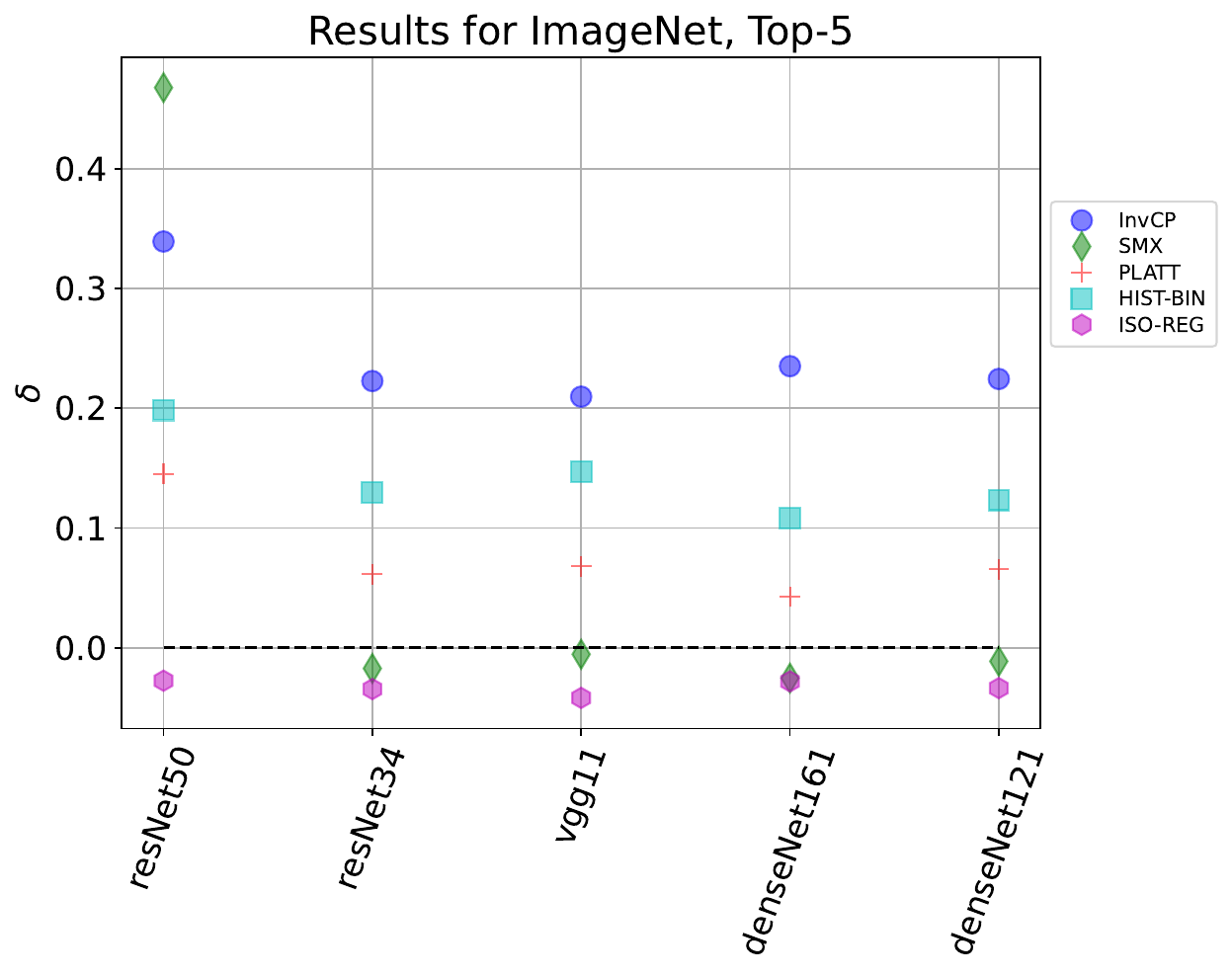}
         \caption{Top-5 results}
         \label{ImageNet_5}
\end{subfigure}
   \caption{Results for ImageNet--V15. Smaller the $\delta$ (see \ref{def: delta}), more accurate the method. A method with $\delta \geq 0$ is conservative.}
   \label{fig: results for ImageNet}
\end{figure}

\textbf{ImageNet-V1:} Results for top-1 model output are shown in \ref{ImageNet_1}. Similar to CIFAR100, the VGG11 model provides the best results with the SMX method. For both the denseNet (161 and 121) and resNet34 models, HIST-BIN provides the best result while being conservative. InvCP fail to provide accurate results for this dataset, despite maintaining conservativeness. The ISO-REG method, similar to CIFAR-100, despite being accurate, fails to maintain conservativeness. Results for top-5 model output are very similar to CIFAR-100---see \ref{ImageNet_5}. Apart from resNet50, SMX provides the best results. The result from ISO-REG is very close to that of SMX.

\begin{figure}[htbp]
\centering
\begin{subfigure}[b]{0.45\textwidth}
\centering
         \includegraphics[width=6cm]{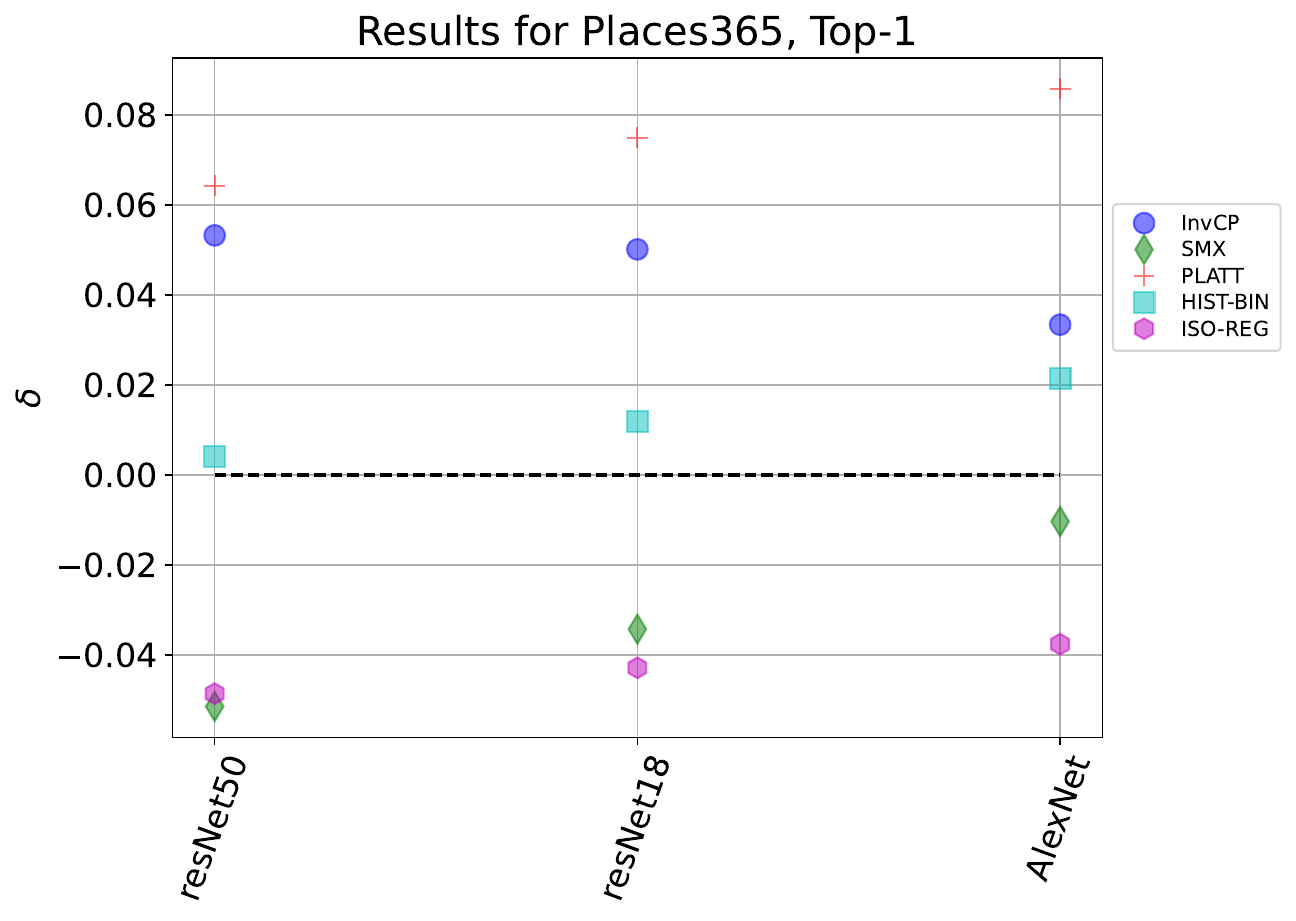}
         \caption{Top-1 results}
         \label{Places365_1}
\end{subfigure}
\begin{subfigure}[b]{0.45\textwidth}
\centering
         \includegraphics[width=6cm]{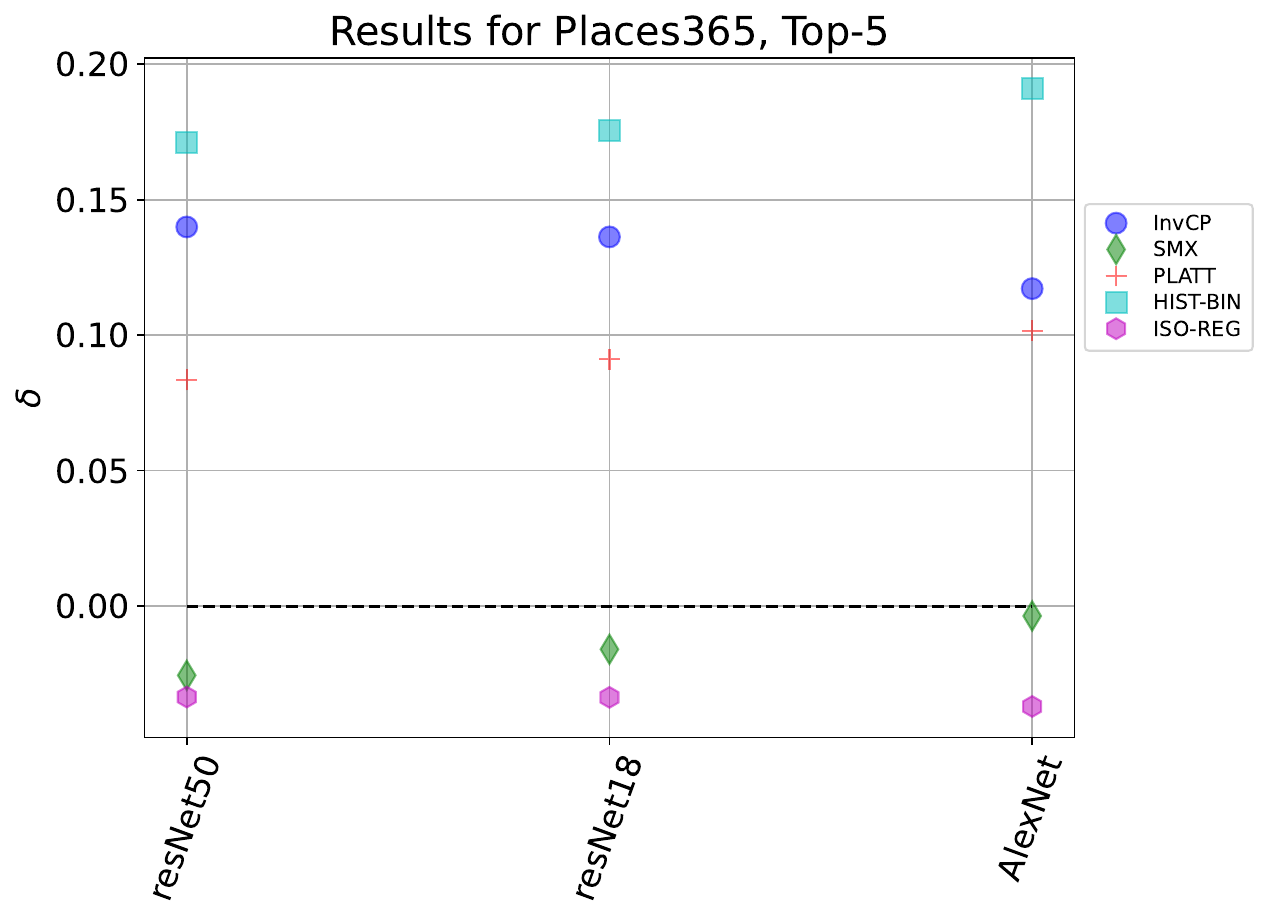}
         \caption{Top-5 results}
         \label{Places365_5}
\end{subfigure}
   \caption{Results for Places365. Smaller the $\delta$ (see \ref{def: delta}), more accurate the method. A method with $\delta \geq 0$ is conservative.}
   \label{fig: results for Places365}
\end{figure}

\textbf{Places365:} Results for top-1 model output are shown in \ref{Places365_1}. For AlexNet, the SMX methods provides the best results. For both the resNet18 and resNet50, HIST-BIN is the most accurate while maintaining conservativeness. For all the models, InvCP is more accurate than the PLATT method and is also conservative. Similar to all the previous datasets, for all of the models, ISO-REG looses conservativeness. Similar to ImageNet, for the top-5 output, SMX provides the best results while ISO-REG results very similar to SMX---see \ref{Places365_5}

\begin{figure}[htbp]
\centering
\begin{subfigure}[b]{0.45\textwidth}
\centering
         \includegraphics[width=6cm]{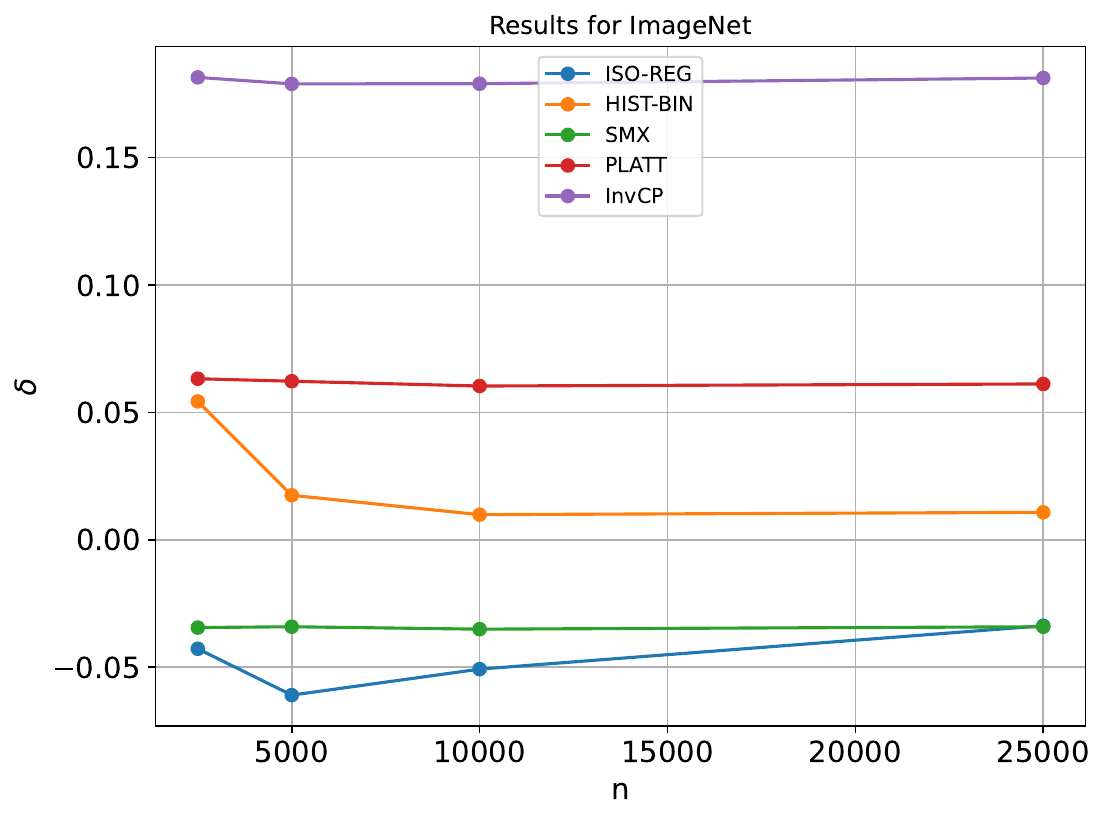}
         \caption{Results for ImageNet}
         \label{diff_n_ImageNet}
\end{subfigure}
\hfill
\begin{subfigure}[b]{0.45\textwidth}
\centering
         \includegraphics[width=6cm]{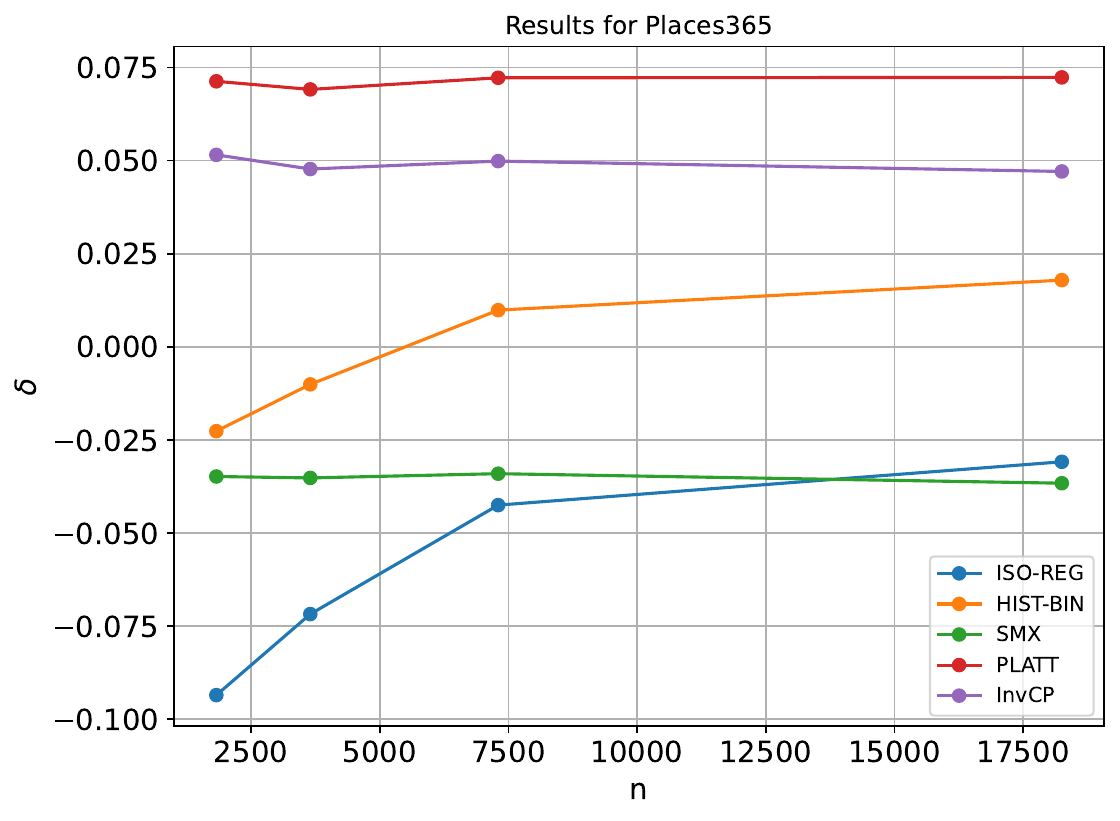}
         \caption{Results for Places365}
         \label{diff_n_Places365}
\end{subfigure}
 \caption{Results for ImageNet--V1 and Places365 for different values of $n$. Computations done with resNet34. \label{diff_n_results}}
\end{figure}

\begin{figure}[htbp]
\centering
         \includegraphics[width=6cm]{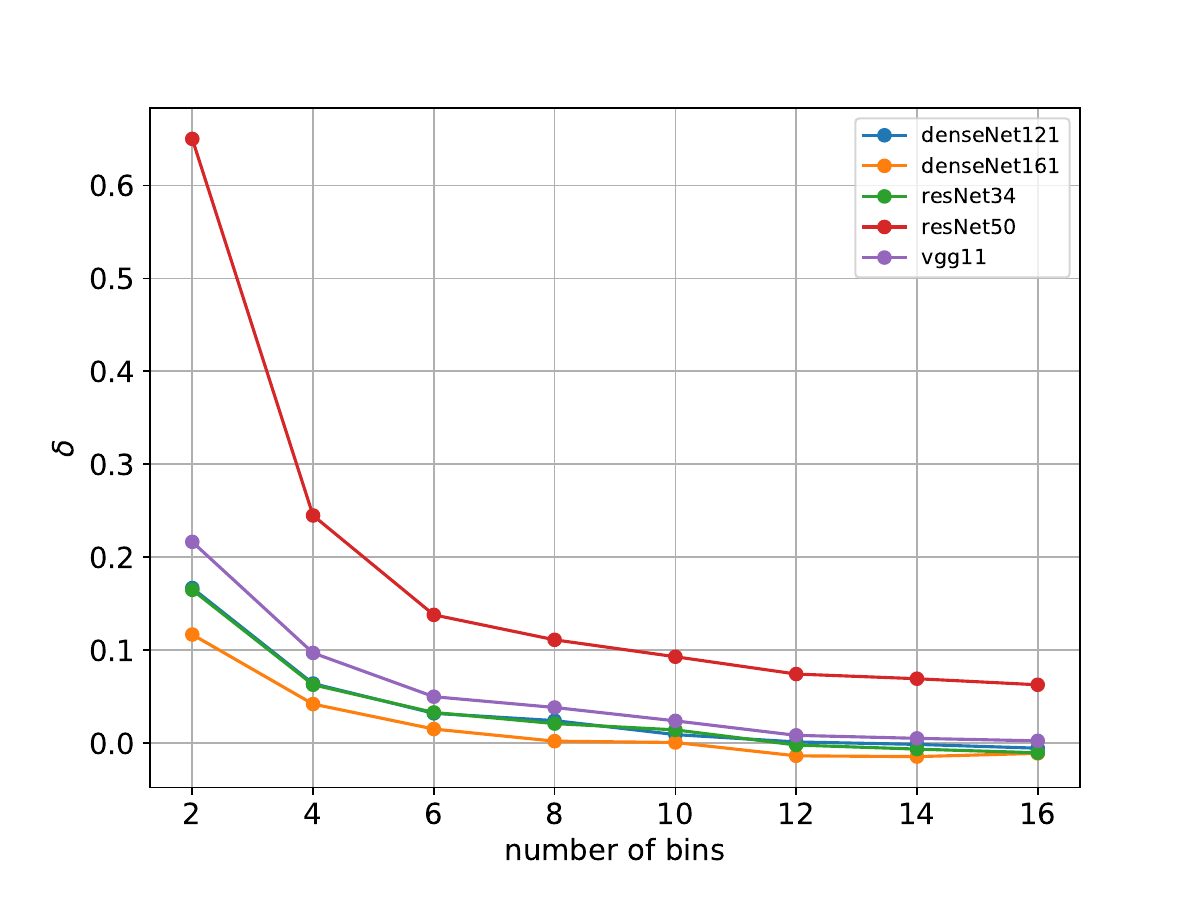}
         \caption{Results for ImageNet for different bin sizes}
         \label{diff_bin_sizes}
\end{figure}

\textbf{Results for different calibration points and bins:} The number of calibration points ($n$) is a parameter for all the methods. For ISO-REG and HIST-BIN, an additional parameter is the number of bins ($M$). We discuss how the results change by varying $n$ and $M$---see \ref{diff_n_results} and \ref{diff_bin_sizes} for the plots. As expected, both ISO-REG and HIST-BIN in general perform better as $M$ increases. Accuracy improvement is drastic as $M$ is increased from $2$ to $8$. A further increase in $M$ does not provide a substantial improvement in accuracy. 

The deviation $\delta$ for both InvCP and PLATT show little fluctuation for different values of $n$. However, $\delta$ for both ISO-REG and HIST-BIN fluctuates substantially as $n$ varies. For instance, consider Places365 for which ISO-REG's accuracy for $n=2500$ is half of that at $n=7500$. Likewise, HIST-BIN looses its conservativeness for small value of $n$. ISO-REG remains non-conservative even for large value of $n$. 

\section{Conclusions and Limitations}
We studied the performance of various methodologies in solving the risk-assessment problem. To this end, standard calibration techniques and a novel conformal prediction based approach were considered. The methodologies were compared in terms of their accuracy and conservativeness i.e., whether the risk was not under-estimated. No particular technique outperformed the others across all model and dataset classes. For datasets with relatively smaller number of labels (CIFAR100, CIFAR10 and Flowers102), the conformal prediction based technique provided in general the best results. For datasets with larger number of labels (ImageNet and Places365), the calibration technique of histogram-binning provided the best results. Despite the shortfall in accuracy, the conformal prediction based technique was conservative throughout. 

The only hyper-parameter in conformal prediction is the size of the calibration set. At least in our experiments, the performance of conformal prediction did not degrade substantially for smaller number of calibration points. In contrast, calibration techniques significantly underperformed with fewer calibration points. Furthermore, the calibration techniques require an appropriate choice of the number of bins otherwise, their performance degrades. No such hyperparameter tuning is required for conformal prediction techniques.

Our investigation offers opportunities for future improvements.
For datasets with large number of labels, the conformal prediction approach could be overly conservative. A detailed theoretical study is required to understand this behaviour better. Furthermore, most practical problems have data-drift and the proposed method needs to be extended to such scenarios.

\bibliography{IEEEexample}
\bibliographystyle{unsrt}
\appendix

\section{Training parameters and accuracies} \label{training for image}
\textbf{Training parameters for CIFAR100, CIFAR10 and Flowers102:} We pre-train all the models for CIFAR100, CIFAR100, CIFAR10 and Flowers102. For the tree-based models, we use the standard functions from sklearn. For logistic regression, the maximum iteration parameter is set to $10000$. We train the CNNs in Pytorch using stochastic gradient descent. The learning rate is set to 0.01, weight decay parameter is set to 1e-4 and the momentum parameter is set to 0.9. We consider a batch size of 256. All the models were trained for 100 epochs. The results reported here correspond to the epoch that resulted in the minimum error rate.

\textbf{Accuracy of different models:} is shown in \ref{table: accuracy} and \ref{table: accuracy2}.

\begin{table}[H]
\centering
\begin{adjustbox}{width=\columnwidth}
\begin{tabular}{ |c|c|c|c|c|c|c| } 
 \hline
  & \multicolumn{3}{|c|}{top-1} & \multicolumn{3}{|c|}{top-5}\\ 
  \hline
  Model & CIFAR100 & ImageNet & Places365 & CIFAR100 & ImageNet & Places365\\
 \hline
 AdaBoost & 7.12 &-&- & 24.8&-&-\\ 
 \hline
 LigthGBM & 19.61&-&- & 47.56&-&- \\
 \hline
 logistic regression & 15.25&-&- & 34.8&-&- \\
 \hline
 random forest & 22.52&-&- & 42&-&-\\
 \hline
 XGBoost & 25&-&- & 50.9&-&- \\
 \hline
 DenseNet121 & 50.7&74.43&- & 81.1&91.97&-\\
 \hline
 DenseNet161 & -&77.13&- & -&93.56&-\\
 \hline
 resNet18 & 46.33&-&53.6 & 77.6&-&83.7\\
 \hline
 resNet34 & 47.61&73.3&- & 79.4&91.4&-\\
 \hline
 resNet50 & 51.26&76.13&54 & 81.6&92.86&84.9\\
 \hline
 VGG11 & 43.62&69&- & 76.3&88.62&-\\
 \hline
 VGG16 & 47.94&-&- & 80.2&-&-\\
 \hline
 AlexNet & -&-& 47.4 & - & -& 77.9\\
 \hline
\end{tabular}
\end{adjustbox}
\caption{Accuracy of different models for different datasets.}
\label{table: accuracy}
\end{table}

\begin{table}[H]
\centering
\begin{adjustbox}{width=\columnwidth}
\begin{tabular}{ |c|c|c|c|c| } 
 \hline
  & \multicolumn{2}{|c|}{top-1} & \multicolumn{2}{|c|}{top-5}\\ 
  \hline
  Model & CIFAR10 & Flowers102 & CIFAR10 & Flowers102\\
 \hline
 AdaBoost & 31.08 & 2.1 & 77.2 & 4.8 \\ 
 \hline
 LigthGBM & 53.1 & 12.2 & 96.3 &  20.0\\
 \hline
 logistic regression  & 37.5 & 18.2 & 77.5 & 71.8 \\
 \hline
 random forest & 47.2 & 15.6 & 98.2 & 83.2\\
 \hline
 XGBoost & 53.8 & 12.9 & 94.3&  94.6\\
 \hline
 DenseNet121 & 86.9 & 41.7 & 97.5 & 96.4\\
 \hline
 resNet18 & 84.6 & 34.01  & 95.5 & 95.1\\
 \hline
 resNet34 & 84.9 & 35.98  & 95.8 & 95.6\\
 \hline
 resNet50 & 87.8 & 31.07  & 98.6 & 97.7\\
 \hline
 VGG11 & 85.3  & 42.3 &  96.2& 96.1\\
 \hline
 VGG16 & 86.1 & 43.6 & 97.3 & 96.5\\
 \hline
\end{tabular}
\end{adjustbox}
\caption{Accuracy of different models for different datasets.}
\label{table: accuracy2}
\end{table}

\begin{figure}[htbp]
\centering
\begin{subfigure}[b]{0.45\textwidth}
\centering
         \includegraphics[width=6cm]{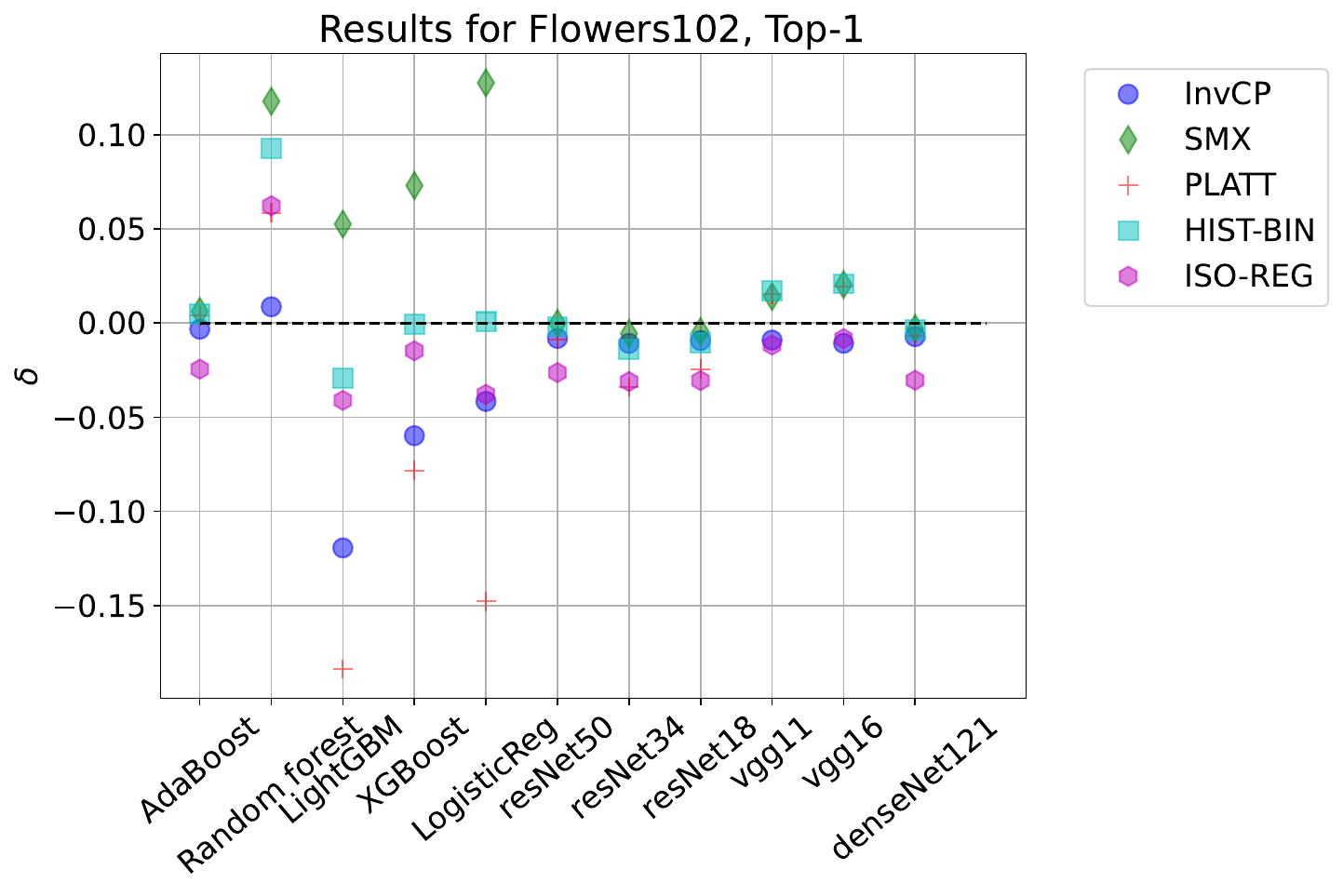}
         \caption{Results for Flowers102, $k=1$}
         \label{Flowers102_1}
\end{subfigure}
\hfill
\begin{subfigure}[b]{0.45\textwidth}
\centering
         \includegraphics[width=6cm]{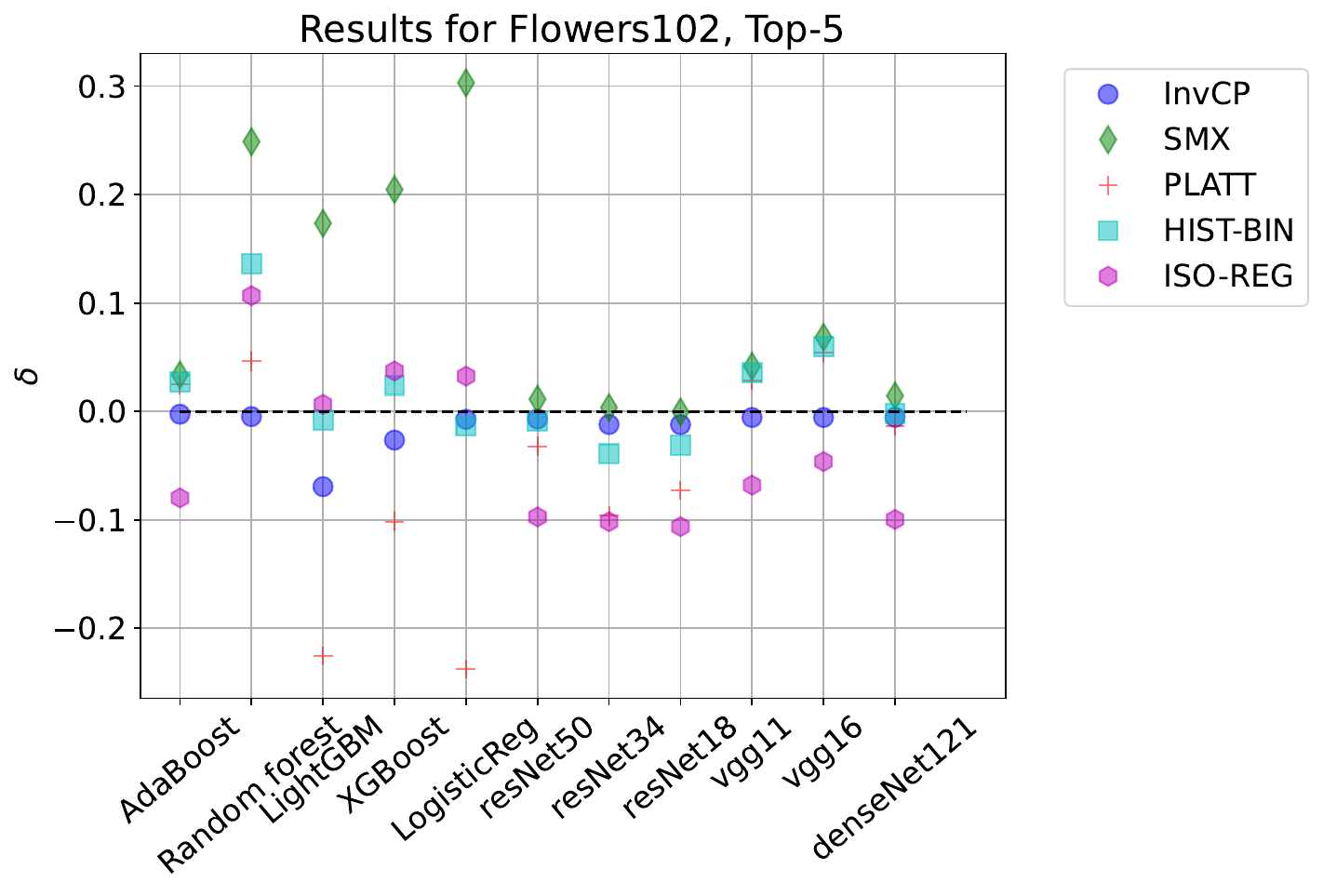}
         \caption{Results for Flowers102, $k=5$}
         \label{Flowers102_2}
\end{subfigure}
 \caption{Results for CIFAR--10 for different values of $k$. \label{fig: Flowers102}}
\end{figure}

\textbf{Results for CIFAR10 and Flowers102}\label{app: further results}
Results for Flowers102 and $k=1$, are shown in \ref{Flowers102_1}. Overall, InvCP provides good results. For AdaBoost, random forest and VGG, it is the best performing model. Compared to HIST-BIN and ISO-REG, it doesn't perform well for the LightGBM model. Nonetheless, it still outperforms Platt scaling. For $k=5$, apart from the LightGBM model, InvCP outperforms all the other methods. 

\begin{figure}[htbp]
\centering
\begin{subfigure}[b]{0.45\textwidth}
\centering
         \includegraphics[width=6cm]{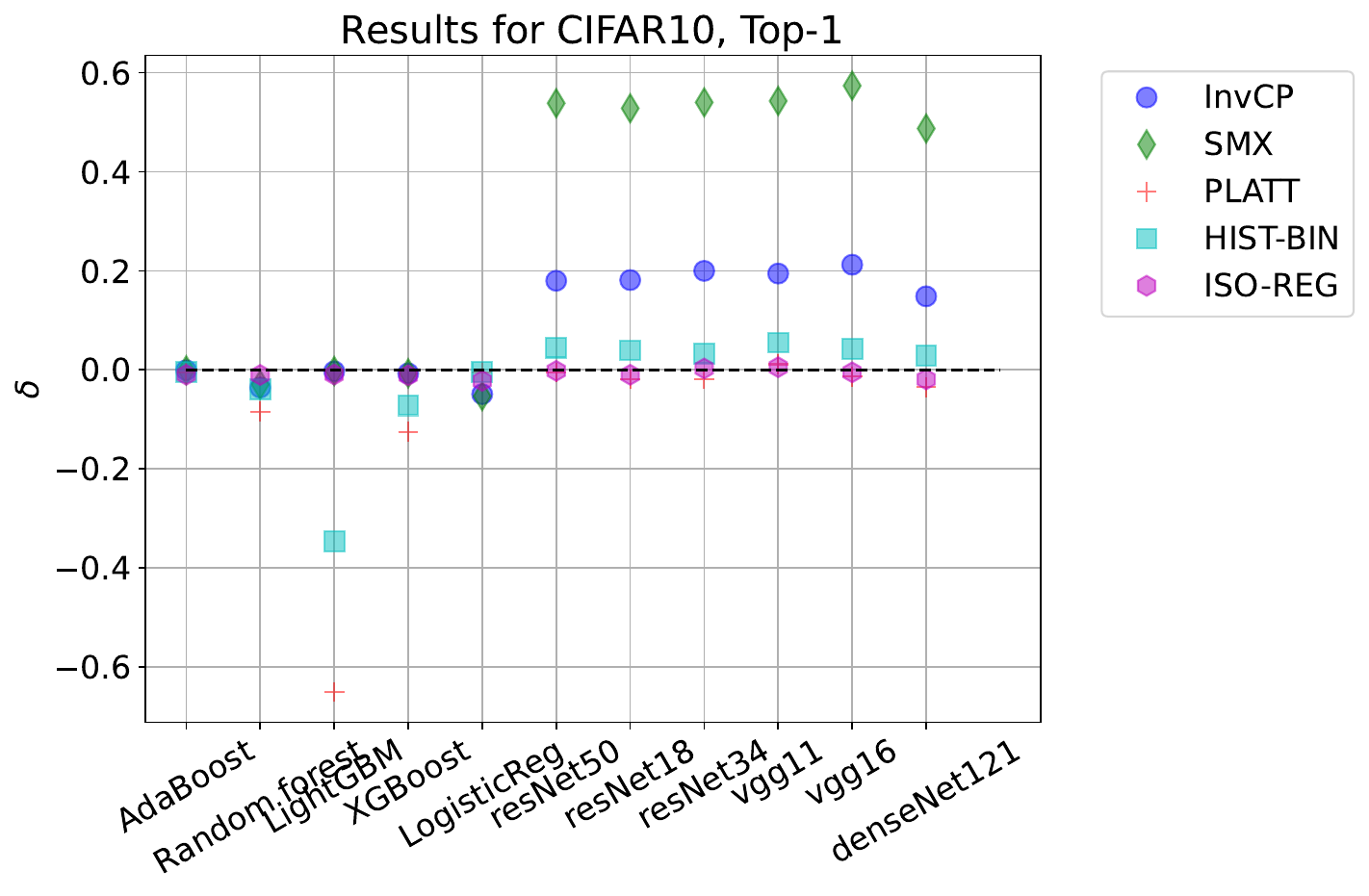}
         \caption{Results for CIFAR10, $k=1$}
         \label{CIFAR10_1}
\end{subfigure}
\hfill
\begin{subfigure}[b]{0.45\textwidth}
\centering
         \includegraphics[width=6cm]{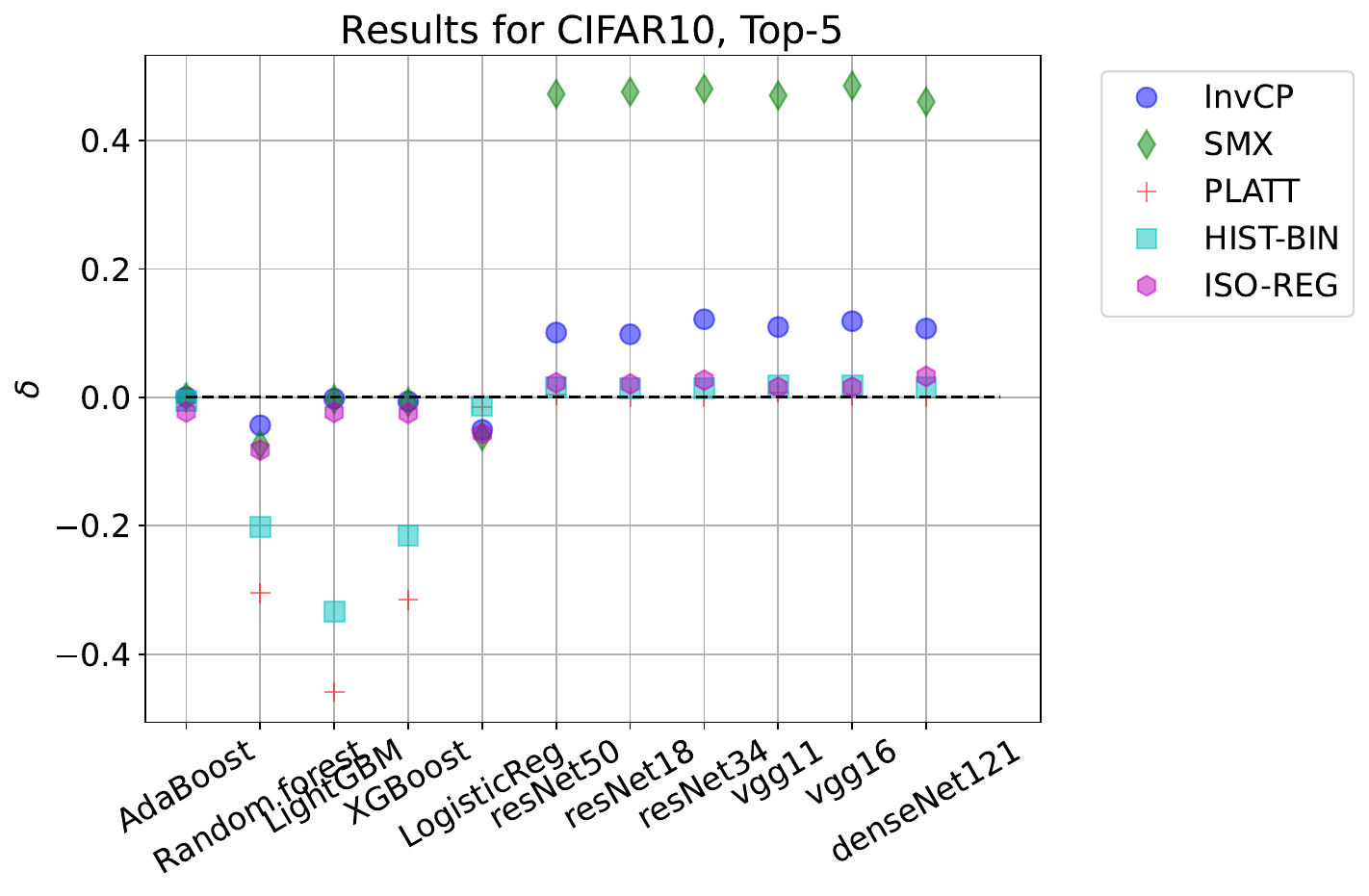}
         \caption{Results for CIFAR10, $k=5$}
         \label{CIFAR10_2}
\end{subfigure}
 \caption{Results for CIFAR--10 for different values of $k$. \label{fig: CIFAR10}}
\end{figure}

Results for CIFAR-10 are presented in \ref{fig: CIFAR10}. For all the tree-based models, the InvCP approach provides the best results. However, for CNNs, InvCP is not the most accurate. Nonetheless, it maintains conservativeness throughout and performs substantially better than the softmax output.

\end{document}